\def\year{2017}\relax
\documentclass[letterpaper]{article} 
\usepackage{ijcai19}  
\usepackage{times}  
\usepackage{helvet}  
\usepackage{courier}  
\usepackage{url}  
\usepackage{graphicx}  
\usepackage[dvipsnames]{xcolor}
\usepackage{amsopn}
\usepackage{amsmath}
\usepackage{amsfonts}
\usepackage{amssymb}
\usepackage{algorithm}
\usepackage[noend]{algpseudocode}
\usepackage{subcaption}
\usepackage{booktabs}
\usepackage{multirow}
\usepackage{amsthm}
\usepackage{textcomp}
\usepackage{listings}%
\usepackage{bm}

\theoremstyle{definition}
\newtheorem{defn}{Definition} 
\newtheorem{prop}{Proposition} 

\makeatletter
\def\BState{\State\hskip-\ALG@thistlm}
\makeatother

\begin{document}
%
\newcommand{\sscomment}[1]{\textcolor{red}{[Sid] #1}}
\newcommand*{\affaddr}[1]{#1} 
\newcommand*{\affmark}[1][*]{\textsuperscript{#1}}
\title{Why Couldn't You do that? \\Explaining Unsolvability of Classical Planning Problems\\ in the Presence of Plan Advice}
%
%
%
%
%
%
%
%

\author{
Sarath Sreedharan\affmark[1],
Siddharth Srivastava\affmark[1],
David Smith\affmark[2]
\and Subbarao Kambhampati\affmark[1]\\
\affaddr{\affmark[1]CIDSE,
Arizona State University, Tempe, AZ 85281 USA}\\
ssreedh3@asu.edu, siddharths@asu.edu,david.smith@psresearch.xyz, rao@asu.edu
}
\maketitle
\begin{abstract}
Explainable planning is widely accepted as a prerequisite for
autonomous agents to successfully work with humans. While there has
been a lot of research on generating explanations of solutions to
planning problems, explaining the absence of solutions remains an open
and under-studied problem, even though such situations can be the
hardest to understand or debug. In this paper, we show that
hierarchical abstractions can be used to efficiently generate reasons
for unsolvability of planning problems. In contrast to related work on
computing certificates of unsolvability, we show that these methods
can generate compact, human-understandable reasons for unsolvability.
Empirical analysis and user studies show the validity
of our methods  as well as their computational efficacy on a number of benchmark
planning domains.
\end{abstract}

\section{Introduction}

The ability to explain the rationale behind a decision is widely seen as one of the basic skills needed by an autonomous agent to truly collaborate with humans.
At the very least we would want our autonomous teammates to be capable of explaining why a particular action/plan was chosen to achieve some objective and be able to explain why they consider some objectives to be unachievable.
For example, consider an automated taxi scheduling system.
A user asks for a taxi to pick up her and three of her friends and the service comes back by saying that it is not possible, and recommends instead using two different taxis.
In this scenario, the user would want to know why a single taxi can't pick up all four of them.

Most earlier works in explanation generation for planning problems have focused on the problem of explaining why a given plan or action was chosen, but do not address the problem of explaining the unsolvability of a given planning problem. While works like \cite{eriksson2018proof,eriksson2017unsolvability} have looked at the problem of generating certificates or proofs of unsolvability, these certificates are geared towards automatic verification rather than human understandability.

In this paper, we present a new approach for explaining unsolvability of planning problems that builds on the well known psychological insight that humans tend to decompose sequential planning problems in terms of the subgoals they need to achieve \cite{donnarumma2016problem,cooper2006hierarchical,simon1971human}. 
We will thus help the user understand the infeasibility of a given planning problem by pointing out unreachable but necessary subgoals. For example, in the earlier case, ``Holding three passengers" is a subgoal that is required to reach the goal, but one that can no longer be achieved due to new city regulations.
Thus the system could explain that the taxi can't hold more than two passenger at a time (and also notify the user about the new city ordinance). 

Unfortunately, this is not so straightforward, since by the very nature of the problem, there exist no solutions and hence no direct way of extracting meaningful subgoals for the problem. We can find a way around this issue by noting the fact that the user is asking for an explanation for unsolvability either due to a lack of understanding of the task or because of limitations in their inferential capabilities. Therefore, we can try to capture the user's expectations by considering abstractions of the given problem. In particular, we use state abstractions to generate potential solutions and subgoals at higher levels of abstractions.
Such an approach was used by \cite{abs-ijcai} to compute explanations for user queries attuned to the level of expertise of the user.

In section \ref{formulation}, we  present our basic framework and discuss how we can identify the appropriate level of abstraction and unachievable subgoals for an unsolvable classical planning problem. 
In the real world, a more challenging version of this problem arises when the user provides \textbf{plan advice} (which may include temporal preferences) on the type of solutions expected. In section \ref{constrained}, we will see how explaining unsolvability of planning problems with plan advice (c.f \cite{myers1996advisable}) could be seen as establishing unsolvability of planning problems with additional plan constraints.
This is a capability that is necessary to capture the fact that these explanations are being provided within the context of a conversation. 
The presence of these additional plan advice could either reflect cases (1) where the original problem was solvable, but users requirements (i.e. expressed in the advice) renders it unsolvable and (2) where the original problem was unsolvable and the user presents an outline for a solution in the form of advice.  Even in the second case, by taking into account the human's expected solution, we can provide a more targeted explanation.
For evaluating our approach, we will look at a user study we ran to validate the usefulness of such explanations for unsolvable problems (with plan advice) and also note the computational efficiency of our method for some standard planning benchmarks.
\begin{figure}[tbp]
\centering
\includegraphics[width=\columnwidth]{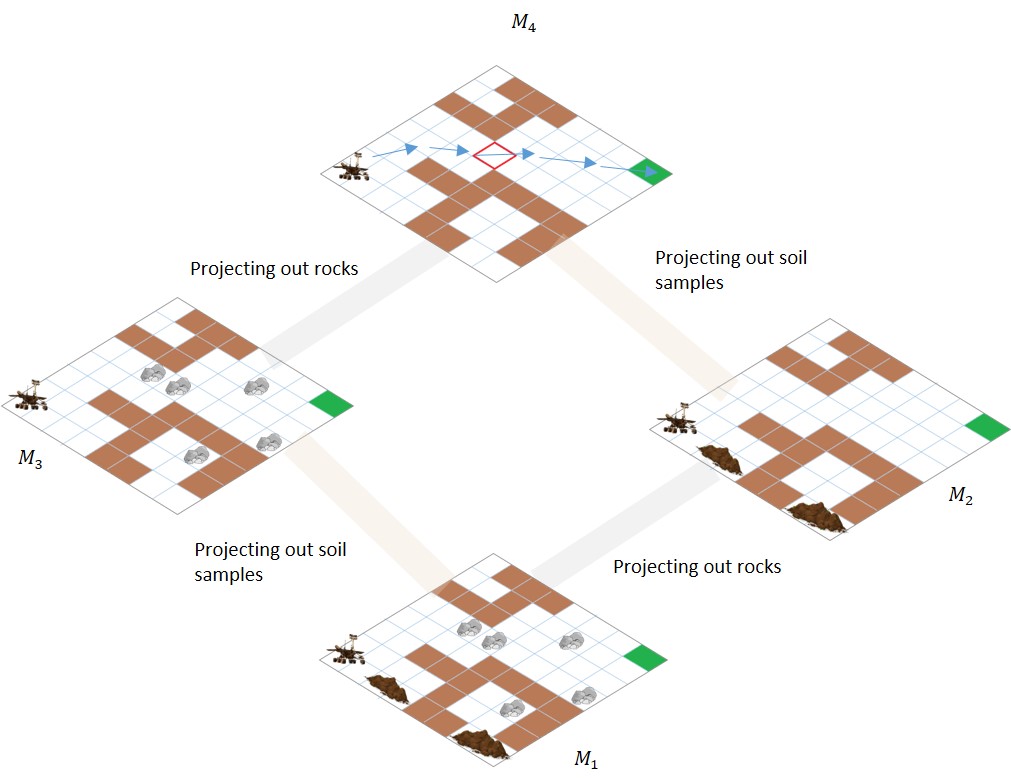}
\caption{{\small A sample abstraction lattice. The lattice consists of models generated by projecting out rocks or soil samples.
The problem is unsolvable (the goal is marked in green) in the most concrete model but solvable in models where the rocks are projected out.}}
\label{fig1}
\vspace{-14pt}
\end{figure}
\section{Background}
We will assume that the autonomous agent uses a STRIPS planning model \cite{fikes1971strips} that can be represented as a tuple of the form $\mathcal{M} = \langle F, A, I, G\rangle$, where $F$ is a set of propositional fluents that define the state space $\mathbb{S}_{M}$ for the model, $A$ gives the set of actions the robot has access to, $I$ defines the initial state and $G$ the goal.
A state $S \in \mathbb{S}_{\mathcal{M}}$ corresponds to a unique value assignment for each state fluent and can be  represented by the set of fluents that are true in that state.
Each action $a \in A$ is further defined by a tuple $a = \langle \textrm{prec}^{a}, \textrm{adds}^{a}, \textrm{dels}^{a}\rangle$ and a plan is defined as an action sequence of the form $\pi = \langle a_1,...,a_n\rangle$. A plan is said to be valid for $\mathcal{M}$, if the result of executing a plan from the initial state satisfies the goal (denoted as $\pi(I) \models_{\mathcal{M}} G $). For the model $\mathcal{M}$, we will represent the set of all valid plans as $\Pi_{\mathcal{M}}$.
Each planning model $\mathcal{M}$ also corresponds to a transition system $\mathcal{T} = \langle \mathbb{S}_{\mathcal{M}}, I, \mathbb{S}_G, A, T\rangle$, where $\mathbb{S}_{G}$ is the subset of $\mathbb{S}_{\mathcal{M}}$ where the goal $G$ is satisfied and $T \subseteq \mathbb{S}_{\mathcal{M}} \times A \times \mathbb{S}_{\mathcal{M}}$, such that $\langle S,a,S'\rangle \in T$ (denoted as $S \xrightarrow[]{a} S'$) if $a(S) = S'$. Each valid plan has a corresponding path in the transition system from I to some state in $\mathbb{S}_G$.

In this work, we will be focusing on state and action abstractions induced by projecting out fluents. 
Thus a model $\mathcal{M}_2$ is said to be an abstraction of $\mathcal{M}_1$ (denoted by $\mathcal{M}_1 \sqsubseteq \mathcal{M}_2$) if model $\mathcal{M}_2$ can be formed from $\mathcal{M}_1$ by projecting out a set of fluents. 
Formally, $\mathcal{M}_1 \sqsubseteq \mathcal{M}_2$ if there exists some $P\subseteq F$, such that the transition system of $\mathcal{M}_2$ is defined as $\mathcal{T}_2 = \langle \mathbb{S}_{\mathcal{M}_2}, I_2, \mathbb{S}_{G_2}, A, T_2\rangle$. 
Where, for every $S \in \mathbb{S}_{\mathcal{M}_1}$, there exist a state $S\setminus P \in \mathbb{S}_{\mathcal{M}_2}$, $I_2 = I \setminus P$, $\mathbb{S}_{G_2}$ is the subset of $\mathbb{S}_{\mathcal{M}_2}$ that satisfy $G' = G \setminus P$ and for every transition $\langle S,a,S'\rangle \in T_1$, there exist $\langle S \setminus P,a,S'\setminus  P\rangle \in T_2$. 
We will denote an abstraction formed by projecting out $P$ from the model $\mathcal{M}$ as $f_P(\mathcal{M})$. 
An abstraction $f_P(\mathcal{M})$ is considered logically complete if for every $\pi$ such that $\pi(I) \models_{\mathcal{M}} G$, we have $\pi(I_{f_P(\mathcal{M})}) \models_{f_P(\mathcal{M})} G_{f_P(\mathcal{M})}$. 
In this work, we will only be looking at logically complete abstractions. For classical planning models, logically complete abstractions can be formed by simply removing the abstracted out fluents from the domain model and problem.

\citeauthor{abs-ijcai} (\citeyear{abs-ijcai}) notes that given a model $\mathcal{M}$ and a set of propositions $P$ we can define an abstraction lattice, denoted as $\mathbb{L} = \langle \mathbb{M}, \mathbb{E}, \ell \rangle$, where each model in $\mathbb{M}$ is an abstraction of  $\mathcal{M}$. 
There exist an edge  $\langle\mathcal{M}_1, \mathcal{M}_2\rangle \in \mathbb{E}$ with label $\ell(\mathcal{M}_1, \mathcal{M}_2) = p$, if $f_{\{p\}}(\mathcal{M}_1) = \mathcal{M}_2$. 
For convenience, we will treat the abstraction function $f$ for a given lattice as invertible and use $f_{P}^{-1}(\mathcal{M})$ to represent the unique concrete node in the lattice that could have been abstracted (by projecting out $P$) to generate $\mathcal{M}$. We will refer to  $f_{P}^{-1}(\mathcal{M})$ as the concretization of $\mathcal{M}$ for $P$. Figure \ref{fig1} presents a simple conceptualization of an abstraction lattice for the rover domain. The edges in the lattice correspond to projecting out the presence of rocks or soil samples.

These earlier work uses such abstraction lattices to estimate the user's level of understanding of the given task, by searching for the level of abstraction where an incorrect alternative raised by the user (or foil) could be supported. In the following section, we will layout our framework and discuss how we leverage the abstraction lattice for our purposes.

\section{Our Approach}
\label{formulation}
The input to our approach includes an unsolvable problem $\mathcal{M}_R = \langle F_R, A_R, I_R, G_R\rangle$ (i.e $|\Pi_{\mathcal{M}_R}| = 0$) and an abstraction lattice $\mathbb{L} = \langle \mathbb{M}, \mathbb{E}, \ell\rangle$, where $\mathbb{M}$ represents the space of possible models that could be used to capture the human's understanding of the task. Given this setting, our method for identifying explanations, includes the following steps
\begin{itemize}
    \item Identify the level of abstraction at which the explanation should be provided (Section \ref{concret})
    \item Identify a sequence of necessary subgoals for the given problem that can be reasoned about at the identified level of abstraction (Section \ref{inferen})
    \item Identify the first unachievable subgoal in that sequence (Section \ref{identify-failed-subgoal})
\end{itemize}
Intuitively, one could understand the three steps mentioned above as follows. First, identify the level of detail at which unsolvability of the problem needs to be discussed. The higher the level of abstraction, the easier the user would find it to understand and reason about the task, but the level of abstraction should be detailed enough that the problem is actually unsolvable there. In most cases, this would mean finding the highest level of abstraction where the problem is still unsolvable.

Now even if the system was to present the problem at this desired level of abstraction, the user may be unable to grasp the reason for unsolvability. Again, our method involves helping the human in this process by pointing out a necessary subgoal (i.e., any valid solution to that problem must achieve the subgoal) that can't be achieved at the current abstraction level. Thus the second point relates to the challenge of finding a sequence of subgoals (defined by state fluents present at the explanatory level) for a given problem. In the third step, we try to identify the first subgoal in the sequence that is actually unsolvable in the given level.

Given our approach, the final explanatory message provided to the user would include model information that brings their understanding of the task to the required level and information on the specific subgoals (and previous ones that need to be achieved first) that can no longer be achieved.
In cases where the unachievable subgoals are hard to understand formulas or large disjunctions, we can also use these subgoals to produce exemplar plans and illustrate their failures alongside the unachievable subgoals. 
\subsection{Identifying the Minimal Level of Abstraction Required for Explanation}
\label{concret}
Let's assume that the human's understanding of the task could be approximated by a model $\mathcal{M}_H = \langle F_H, A_H, I_H, G_H\rangle$, such that, the model is part of the abstraction lattice ($\mathcal{M}_R \sqsubset \mathcal{M}_H$ and $ \mathcal{M}_H \in \mathbb{M}$) and since the user expected the problem to be solvable, $\mathcal{M}_H$ is such that  $\exists \pi, \pi (I_{\mathcal{M}_H}) \models_{\mathcal{M}_H} G_{\mathcal{M}_H}$, i.e $\pi \in \Pi_{\mathcal{M}_H}$.

We now need to this human model to an abstraction level where the problem is unsolvable (i.e the explanation level) by providing information about a certain subset of fluents previously missing from the human model (i.e information on their truth values in the initial and goal state, and how they affect various actions etc...).

For example, in the case of Figure \ref{fig1}, let us assume that the human model is $\mathcal{M}_2$, then the information that needs to be provided to the user involves the position of the rocks and how they restrict certain robot motions. 
We will refer to the set of fluents that the human needs to be informed about as explanatory fluents ($\mathcal{E}$) and for Figure \ref{fig1}, it will be  $\mathcal{E} = \{has\_rocks(?x,?y)\}$.

\begin{defn}
{\em Given a human model $\mathcal{M}_H$, we define a set of propositions $\mathcal{E}$ to be \textbf{explanatory fluents} if $f_{\mathcal{E}}^{-1}(\mathcal{M}_H)$ is unsolvable, i.e, $|\Pi_{f_{\mathcal{E}}^{-1}(\mathcal{M}_H)}| = 0$.}
\end{defn}
Unfortunately, this is not an operational definition as we do not have access to $\mathcal{M}_H$.
Instead, we know that $\mathcal{M}_H$ must be part of the lattice, and thus there exists a subset of the maximal elements of the lattice (denoted as $\mathbb{M}^{abs}$)\footnote{w.l.o.g we assume the existence of a set of maximal elements instead of a unique supremum as these lattices need not be complete.} that is more abstract than $\mathcal{M}_H$.
In this section, we will show how the explanatory fluents for models in this subset of $\mathbb{M}^{abs}$ would satisfy $\mathcal{M}_H$ as well.

The first useful property to keep in mind is that if $\mathcal{M}_1$ is more concrete than $\mathcal{M}_2$ then the models obtained by concretizing each model with the same set of fluents would maintain this relation (although they may get concretized to the same model), i.e.,
\begin{prop}
{\em Given models $\mathcal{M}_1$, $\mathcal{M}_2$ and a set of fluents $\epsilon'$, if $\mathcal{M}_1 \sqsubseteq \mathcal{M}_2$, then $f^{-1}_{\epsilon'}(\mathcal{M}_1) \sqsubseteq f^{-1}_{\epsilon'}(\mathcal{M}_2)$ .} 
\end{prop}

Next, it can be shown that any given set of explanatory fluents for an abstract model will be a valid explanatory fluent set for a more concrete model

\begin{prop}
\label{resolv-abs}
{\em Given models $\mathcal{M}_1$, $\mathcal{M}_2$, if $\mathcal{M}_1 \sqsubseteq \mathcal{M}_2$, then any explanation $\mathcal{E}$ for $\mathcal{M}_2$ must also be an explanation for $\mathcal{M}_1$.}
\end{prop}
To see why this proposition is true, let's start from the fact that $f^{-1}_{\mathcal{E}}(\mathcal{M}_1)\sqsubseteq f^{-1}_{\mathcal{E}}(\mathcal{M}_2)$ and therefore $\Pi_{f^{-1}_{\mathcal{E}}(\mathcal{M}_1)} \subseteq \Pi_{f^{-1}_{\mathcal{E}}(\mathcal{M}_2)}$. 
From the definition of explanation we know that the concretization with respect to explanatory fluents would render the problem unsolvable (i.e $|\Pi_{f^{-1}_{\mathcal{E}}(\mathcal{M}_2)}| = 0$) and thus $|\Pi_{f^{-1}_{\mathcal{E}}(\mathcal{M}_1))}|$ must also be empty and hence $\mathcal{E}$ is an explanation for $\mathcal{M}_1$.

\begin{defn}
{\em Given an abstraction lattice $\mathbb{L}$, let $\mathbb{M}^{abs}$ be its maximal elements. Then the \textbf{minimum abstraction set} is defined as $\mathbb{M}_{min} = \{\mathcal{M}| \mathcal{M} \in \mathbb{M}^{abs} \wedge |\Pi_{\mathcal{M}}| > 0 \}$.}
\end{defn}

Note that for any model $\mathcal{M}_1 \in \mathbb{M}_{min}$, $\mathcal{M}_H \sqsubseteq \mathcal{M}_1$, this means by Proposition \ref{resolv-abs}, any explanation that is valid for models in $\mathbb{M}_{min}$, should lead $\mathcal{M}_H$ to a node where the problem is unsolvable. Now we can generate the explanation (even the optimal one) by searching for a set of fluents that when introduced to the models $\mathcal{M} \in \mathbb{M}_{min}$ will render resulting $f^{-1}_{\mathcal{E}}(\mathcal{M})$ unsolvable. 

\subsection{Generating Subgoals of a Given Problem}
\label{inferen}
Note that we can't identify possible subgoals for the given problem in the node at which the problem was found to be unsolvable (i.e $f_{\mathcal{E}}^{-1}(\mathbb{M}_{min})$). 
There exist no valid plans in that model and thus there are no subgoals to point to. 
Fortunately, we can use models more abstract than $f_{\mathcal{E}}^{-1}(\mathbb{M}_{min})$ to generate such subgoals. 
We will use planning landmarks \cite{hoffmann2004ordered} extracted from $\mathcal{M}$, where $|\Pi_{\mathcal{M}}| > 0$, as subgoals. 
Intuitively, state landmarks (denoted as $\Lambda = \langle \Phi, \prec\rangle$) for a model $\mathcal{M}$ can be thought of as a partially ordered set of formulas, where the formulas and the ordering needs to be satisfied by every plan that is valid in $\mathcal{M}$. 
We will only be considering sound orderings (c.f \cite{richter2008landmarks}) between landmarks, namely, (1) \textit{natural orderings} ($\prec_{nat}$) - $\phi \prec_{nat} \phi'$, then $\phi$ must be true before $\phi'$ is made true in every plan, (2) \textit{necessary orderings} ($\prec_{nec}$) - if $\phi \prec_{nec} \phi'$ then $\phi$ must be true in the step before $\phi'$ is made true every time and (3) \textit{greedy necessary orderings} ($\prec_{gnec}$) - if $\phi \prec_{gnec} \phi'$ then $\phi$ must be true in the step before $\phi'$ is made true the first time. 
The landmark formulas may be disjunctive, conjunctive or atomic landmarks. 

Our use of landmarks as the way to identify subgoals is further justified by the fact that logically complete abstractions conserve landmarks. Formally
\begin{prop}
\label{abs-land}
{\em Given two models $\mathcal{M}_1$ and $\mathcal{M}_2$, such that  $\mathcal{M}_1 \sqsubseteq \mathcal{M}_2$, let $\Lambda_1=\langle \Phi_1, \prec_1\rangle$ and $\Lambda_2=\langle \Phi_2, \prec_2\rangle$ be the landmarks of $\mathcal{M}_1$ and $\mathcal{M}_2$ respectively. Then for all $\phi_i^1, \phi_j^1 \in \Phi_1$, such that $\phi_i^1 \preceq_1 \phi_j^1$ (where $\prec_1$ is some sound ordering), we have $\phi_i^2$ and $\phi_j^2$ in  $\Phi_2$, where $\phi_i^1 \preceq_2 \phi_j^1$, $\phi_i^1 \models \phi_i^2$ and $\phi_j^1 \models \phi_j^2$.}
\end{prop}
This is true because $\phi_i^2 \prec_1 \phi_j^2$ hold over all the plans that are valid in $\mathcal{M}_2$, thus must also hold over all plans in $\mathcal{M}_1$. 
Though in $\mathcal{M}_1$ these landmark instances may be captured by more constrained formulas and additionally $\mathcal{M}_1$ may also contain landmarks that were absent from $\mathcal{M}_2$. 
Now if we can show that in a particular model, a landmark generated from a more abstract model is unachievable (or the  ordering from the previous level is unachievable) then $\phi_*^1$ becomes $\bot$ (thereby meeting the above requirement). 
Thereafter for any model more concrete than $\mathcal{M}_2$, the formula corresponding to that landmark must be $\bot$.
In other words, if for any model a landmark is unachievable, then that landmark can't be achieved in any models more concrete than the current one.

So given the explanatory level, we can move one level up in the lattice and make use of any of the well established landmark extraction methods developed for classical planning problem to generate a sequence of potential subgoals for $\mathcal{M}_R$.

\subsection{Identifying Unachievable Sub-Goals}
\label{identify-failed-subgoal}
Now we need to find the first subgoal from the sequence that can no longer be achieved in the models obtained by applying the explanatory fluents ( $f^{-1}_{\mathcal{E}}(\mathbb{M}_{min})$) which will then be presented to the user.
For example, in the case of Figure \ref{fig1}, the unachievable subgoal would correspond to satisfying $at\_rover(5,4)$ (marked in red in $M_4$).

It is important to note that finding the first unachievable subgoal is not as simple as testing the achievability of each subgoal at the abstraction level identified by methods discussed in section \ref{concret}.
Instead, we need to make sure that each subgoal is achievable while preserving the order of all the previous subgoals. 
To test this we will introduce a new compilation that allows us to express the problem of testing achievement of a landmark formula as a planning problem. 
Consider a planning model $\mathcal{M}$ and the landmarks $\Lambda = \langle \Phi, \prec\rangle$ extracted from some model $\mathcal{M}'$, where $\mathcal{M} \sqsubset \mathcal{M}'$. 
We will assume that the formulas in $\Phi$ are propositional logic formulas over the state fluents and are expressed in DNF. 
Each $\phi \in \Phi$ can be represented as a set of sets of fluents (i.e, $\phi=\{c_1,...,c_k\}$ and each $c_i$ set takes the form $c_i= \{p_1,..p_m\}$), where each set of fluents represent a conjunction over those fluents. 
For testing the achievability of any landmark $\phi \in \Phi$, we make an augmented model $\mathcal{M}_\phi = \langle F^\phi, A^\phi, I^\phi, G^\phi\rangle$, such that the landmark is achievable \textit{iff} $|\Pi_{\mathcal{M}_\phi}| > 0$. The model $\mathcal{M}_\phi$ can be defined as follows:
$F^\phi=F\cup F^{meta}$, where $F^{meta}$ contains new meta fluents for each possible landmark $\phi' \in  \Phi$  of the form 
\begin{itemize}
\item $achieved(\phi')$ keeps track of a landmark being achieved and never gets removed
\item $unset(\phi')$ Says that the landmark has not been achieved yet, usually set true in the initial state unless the landmark is true in the initial state
\item $first\_time\_achieved(\phi')$ Says that the landmark has been achieved for the first time. This fluent is set true in the initial state if the landmark is already true there
\end{itemize}
The new action set $A^\phi$, will contain a copy of each action in $A$. For each new action corresponding to $a \in A$, we add the following new  effects to track the achievement of each landmark 
\begin{itemize}
\item for each $\phi' \in \Phi$ if the action has existing add effects for a subset of predicates $\hat{c}_j$ for a $c_j \in \phi'$, then we add the conditional effects $ cond_1(\phi') \rightarrow \{achieved(\phi')\}$  and $cond_2(\phi') \rightarrow \{first\_time\_achieved(\phi')\}$, where \\$cond_1(\phi') = c_j \setminus \hat{c}_j \cup \{\hat{\phi} | \hat{\phi}\in \Phi \wedge (\hat{\phi} \prec_{nec} \phi')\}  \cup \{achieved(\hat{\phi})|\hat{\phi} \prec_{nat} \phi' \} $ and \\$cond_2(\phi') = cond_1(\phi') \cup \{ \hat{\phi} |\hat{\phi} \prec_{gnec} \phi'\}\cup \{unset(\phi)\}$
\item We add a conditional delete effect to every action of the form $first\_time\_achieved(\phi') \rightarrow (not (first\_time\_achieved(\phi')))$
\end{itemize}
The new goal would be defined as $G^\phi = \{first\_time\_achieved(\phi)\}$.

This formulation allows us to test each landmark in the given sequence and find the first one that can no longer be achieved.
To ensure completeness, we will return the final goal if all the previously extracted landmarks are still achievable in $f^{-1}_{\mathcal{E}}(\mathbb{M}_{min})$. Since the above formulation is designed for DNF, we can generate compilation for cases where the landmarks use either un-normalized formulas or CNF by converting them first into DNF formulas.
Readers can find sample explanations generated using our methods in the supplementary file hosted at \url{https://goo.gl/nc2NP3}.

\section{Planning Problem with Plan Advice}
\label{constrained}
Let us now discuss how we could extend the methods presented in earlier sections to cases where the user provides plan advice. In such cases,the user imposes certain restrictions on the kind of solution they expect, either as an alternative to the solution the system may come up with on its own or as a guide to help the system come up with solutions when it claims unsolvability.

As pointed out in \cite{myers1996advisable}, such advice can be compiled into plan constraints in the original problem.
A number of approaches have been proposed to capture and represent plan constraints \cite{bacchus2000using,nau2001shop,kambhampati1995planning,baier2006planning}, and each of these representational choices has its unique strengths and weaknesses.
In general, we can see that these plan constraints specify a partitioning of the space of all valid plans to either acceptable (i.e it satisfies the constraints) or unacceptable. So we can define, constraints as follows
\begin{defn}
{\em A \textbf{constraint $\sigma$} for a planning model $\mathcal{M}$, specifies a function that maps a set of valid plans on $\mathcal{M}$ to a subset, $\sigma: 2^{\Pi_\mathcal{M}} \rightarrow 2^{\Pi_\mathcal{M}}$, such that, $\sigma(\Pi_\mathcal{M}) \subseteq \Pi$.}
\end{defn}

If we can assume some upper bound on the possible length of plans in $\sigma(\Pi_{\mathcal{M}})$ (which is guaranteed when we restrict our attention to non-redundant plans in any function free fragment of classical planning), then we can assert that there always exists a finite state machine that captures the space of acceptable plans
\begin{prop}
{\em Given a constraint $\sigma$, there exists a finite state automaton $\mathcal{F}^{\sigma} = \langle \Sigma, \mathbb{S}_{\mathcal{F}^{\sigma}}, S_0, \delta, E\rangle$, where $\Sigma$ is the input alphabet, $\mathbb{S}_{\mathcal{F}^{\sigma}}$ defines the FSA states, $S_0$ is the initial state, $\delta$ is the transition function and $E$ is the set of accepting states, such that $\sigma(\Pi_{\mathcal{M}}) = \mathcal{L}(\mathcal{F}^{\sigma}) \cap ~\Pi_{\mathcal{M}}$, where $\mathcal{L}(\mathcal{F}^{\sigma})$ is the set of strings accepted by $\mathcal{F}^{\sigma}$.}
\end{prop}
The existence of $\mathcal{F}^{\sigma}$ can be trivially shown by considering an FSA that has a path for each unique plan in $\mathcal{F}^{\sigma}$. We believe that this formulation is general enough to capture almost all of the plan constraint specifications discussed in the planning literature, including LTL based specifications, since for classical planning problems these formulas are better understood in terms of $LTL_{f}$ \cite{de2015synthesis} which can be compiled into a finite state automaton.

We can use $\mathcal{F}^{\sigma}$ to build a new model $\sigma(\mathcal{M})$ such that a plan is valid in $\sigma(\mathcal{M})$ if and only if the plan is valid for $\mathcal{M}$ and satisfies the given specification $\sigma$, i.e., {\small$ \forall \pi, \pi\in \Pi_{\sigma(\mathcal{M})} \textrm{ \textit{iff} } \pi \in \sigma(\Pi_{\mathcal{M}})$} 

For $\mathcal{M} = \langle F, A, I, G\rangle$, we can define the new model $\sigma(\mathcal{M}) =  \langle F_{\sigma}, A_{\sigma}, I_{\sigma}, G_{\sigma}\rangle$ as follows
{\small
\begin{itemize}
\item $F_{\sigma} = F \cup \{\textrm{in-state-}\{S\}|S \in \mathbb{S}_{\mathcal{F}^\sigma}\}$
\item $A_{\sigma} = A \cup A_{\delta}$
\item $I_{\sigma} = I \cup \{\textrm{in-state-}\{S_0\}\}$
\item $G_{\sigma} = G \cup \{\textrm{in-state-}\{S\}| S\in E\}$
\end{itemize}
}
$A_{\delta}$ are the new meta actions responsible for simulating the transitions defined by $\delta: \mathbb{S}_{\mathcal{F}^\sigma} \times \Sigma \rightarrow pow(\mathbb{S}_{\mathcal{F}^\sigma})$. For example, if $\delta(S_1,a) = \{S_1,S_2\}$, where $a$ corresponds to an action in $A$, then we will have two new actions $a_{S_1,a}^1 = \langle prec^a\cup\{\textrm{in-state-}\{S_1\}\}, adds^a\cup\{\textrm{in-state-}\{S_2\}\}, dels^a\cup\{\textrm{in-state-}\{S_1\}\}\rangle$ and $a_{S_1,a}^2 = \langle prec^a\cup\{\textrm{in-state-}\{S_1\}\}, adds^a, dels^a\}\rangle$. In cases like LTL, the FSA state transitions may be induced by the satisfaction of some formula, so the new meta action may have preconditions corresponding to that formula, with no other effects but changing the fluent corresponding to the state transition.

The above formulation merely points out that there always exists a way of generating $\sigma(\mathcal{M})$ from the given specification $\sigma$ and $\mathcal{M}$. For many constraint types, there may exist more efficient ways of generating models that satisfy the requirements of $\sigma(\mathcal{M})$. 

Once we have access to $\sigma(\mathcal{M})$, we should be able to use the methods discussed in earlier sections, provided we can show that these new models could be used to build an abstraction lattice. In particular, we want to know if the abstraction relations are preserved. Fortunately this is true. Formally,

\begin{prop}
{\em Given models $\mathcal{M}_1$, $\mathcal{M}_2$ and a constraint specification $\sigma$, if $\mathcal{M}_1 \sqsubseteq \mathcal{M}_2$, then $\sigma(\mathcal{M}_1) \sqsubseteq \sigma(\mathcal{M}_2)$.}
\end{prop}
To see why this is true, just assume that the reverse was true, that $\sigma(\mathcal{M}_2)$ is not a logically complete abstraction of $\sigma(\mathcal{M}_1)$. This means that there are plans in $\Pi_{\sigma(\mathcal{M}_1)}$ that are not part of $\Pi_{\sigma(\mathcal{M}_2)}$. From the definition of $\sigma(\mathcal{M}_2)$, we know that $\Pi_{\sigma(\mathcal{M}_2)} = \Pi_{\mathcal{M}_2} \cap \mathcal{L}(\mathcal{F}^\sigma)$. If there exist a $\pi \in \Pi_{\sigma(\mathcal{M}_1)}$, such that  $\pi \not \in \Pi_{\sigma(\mathcal{M}_2)}$, then  $\pi \not \in \Pi_{\mathcal{M}_2}$. Which means $\mathcal{M}_1 \not\sqsubseteq \mathcal{M}_2$, hence contradicting our assumptions.


\begin{figure*}[t!]
\small
  \begin{tabular}{|l|c|c|c|c|c|c|}
    \hline
    \multirow{3}{*}{Domain Name} &
    \multicolumn{3}{c|}{\multirow{1}{*}{Uknown Human Model}} &\multicolumn{3}{c|}{\multirow{1}{*}{Known Human Model}} \\
       &lattice size& Explanation Cost&Average Runtime (secs)&lattice size&  Explanation Cost&Average Runtime (secs)\\
      \hline
      \multirow{1}{*}{Elevator} &8&3.4&0.772&3&3.4&0.529\\
      \hline
      \multirow{1}{*}{Blocksworld} &4&11.6&8.141&2&11.6&14.77\\
      \hline
      \multirow{1}{*}{Satellite} &8&6.5&8.586&4&6.5&5.18\\
            \hline
      \multirow{1}{*}{Depots} &5&13&20.229&3&12&44.920\\
            \hline
      \multirow{1}{*}{Rover} &10&3.8&365.287&5&3.8&338.944\\
      \midrule
  \end{tabular}
\vspace{-5pt}
\caption{
{\small Table showing runtime for explanations generated for standard IPC domains. The explanation costs capture the number of unique model updates (changes in effects/precondition etc..) corresponding to each explanation}}
\label{tab1}
\vspace{-10pt}
\end{figure*}

\section{Evaluations}
\label{eval}
\subsection{User Studies}
Our first topic for evaluation is whether explanations based on landmarks do, in fact, constitute meaningful explanations, at least for naive users. As a simple alternative, we enumerated over a set of solutions (generated from a higher level of abstraction) and pointed out their individual failures. For the study, we recruited around 120 master turkers from Amazon's Mechanical Turk  and tested the following hypotheses
\begin{itemize}
\item \textbf{H1} - Users prefer explanations concise explanations over ones that enumerate a set of possible candidates for a given piece of plan advice
\item \textbf{H2} - Users prefer concise explanations that contain information about unachievable landmarks over ones that only show the failure of a single exemplary plan
\end{itemize}
For the hypotheses, we presented the study participants with a sample dialogue between two people over a logistics plan to move a package from one location to another. The dialogue included a person (named Bob) presenting a plan to another (named Alice), and Alice asks for an alternative possibility (i.e specifies a constraint on the solution). Now the challenge for Bob is to explain why the constrained problem is unsolvable.

For H1, the potential explanations included (a) just information on the unachievable landmark, (b) landmark information with the failure details of a specific exemplary plan and (c) a set of three plans that satisfy the constraints and their corresponding failures. For this study, we used 45 participants and each participant was assigned one of three possible maps for each hypothesis and was paid \$1.25 for 10 mins. We used a control question to filter participant responses, so as to ensure their quality. Out of the 39 remaining responses, we found 94.8\% of users chose to select the more concise explanation (i.e (a) or (b)), and 51.28\% of the users chose explanations that involved just landmarks.

For H2, we used 75 participants and presented each participant with explanations that include (a) just landmark information, (b) landmark information with failure details of an exemplary plan and (c) just the exemplary plan failure. Here participants were paid \$1 for 10 mins for H2 as the explanatory options were much simpler. After filtering using the control question, we found that out of 60 valid entries 75.4\% of participants preferred explanations that included landmark information ((a) or (b)) and 44.2\% wanted both landmarks and exemplary plan (i.e (b)). The supplementary file at \url{https://goo.gl/nc2NP3} contains more details on the study setup.
\subsection{Empirical Studies}
\label{emp-eval}
In this section, we will present the results of an empirical evaluation of the computational charecteristics of our approach. 
One big concern with the methods discussed in this work is the fact that they involve solving multiple planning problems. 
Thus we were interested in identifying how the runtime for an explanation would change when the lattice size changed.
As a baseline for comparison we use the trivialized problem where the agent must compute an explanation for a user whose model is known, which means the corresponding search space is much smaller (the supremum of the lattice will be the unique human model).

We considered five standard IPC domains and generated a stand-in for the human model by projecting out a random subset of state fluents. 
Next, we chose five problem instances for each of the domains and recorded the time required to generate explanations for the human model versus a case where the model was unknown. 
We simulated the latter case by considering a complete abstraction lattice generated from a superset of the fluents that the human is actually unaware of. 
For each of the five domains, the total number of fluents used to generate the lattice for the unknown case was at least twice the original number of fluents that was missing from the human model. 
Each problem instance was made unsolvable by including plan constraints that avoid a specific landmark of the original problem. 
The constraints were coded using domain control programs of the type discussed in \cite{baier2007exploiting}. The constraints ensures that plans avoid one of the landmarks of the original problem, thereby rendering it unsolvable.

Figure \ref{tab1}, shows the average runtime and cost of explanations (the number of model updates corresponding to an explanation) related to each domain. 
We can see that the conciseness of the explanations does not suffer when the user model is not known!
For most domains, the total runtime is quite comparable between the two cases, with the unknown case performing better in some domains. 
This is due to the fact that in the unknown case the test for solvability is usually carried out in more abstract models.
\section{Related Work}
\vspace{-1pt}
As mentioned, there has been prior work on generating planner independent certificates or proofs for establishing the unsolvability of a given planning problem.
Another related direction, has been the effort to generate ``excuses" for unsolvability of a planning problem \cite{gobelbecker2010coming,herzig2014revision}. They do not provide an intuitive explanation as to why a problem is unsolvable, but rather identifies initial state values (or some domain model conditions) whose update could make the problem solvable.

The contrastive explanations of the type studied in \cite{abs-ijcai}, where the user presents an alternative plans (that are then refuted by the system) can be thought of as a special case of our approach for problems with plan advice. The problems studied in that earlier paper can be thought about cases where the advice only allows a single plan. Also, one could argue that people would be more comfortable giving advices than full plans.

Part of our explanations also try to reveal to the user information about the current task that was previously unknown to them.
Thus our methods could also be understood as an example of explanation as model-reconciliation \cite{explain}.
Since our methods use abstractions, our approach doesn't make too many demands on the inferential capabilities of the user and hence can be applied to much larger and more complex domains.

Another closely related direction has been the work done on explaining unsynthesizability of hybrid controllers for a given set of high-level task specifications \cite{raman2013towards}.
The work tries to identify the subformulas of the given specification that lead to the unsynthesizability.
This particular approach is specific to the planning framework detailed in \cite{finucane2010ltlmop} and the objective of the work parallels the goals of work like \cite{gobelbecker2010coming,herzig2014revision}.

Outside of explanation generation, the work done in the model checking community is closely related to our current problem \cite{grumberg200825}. In fact, the hierarchical approach to identifying a model that can invalidate the given foil specification, can be seen as a special case of the CEGAR based methods studied in the model-checking community \cite{clarke2000counterexample}. Most work in this field focuses on developing methods for identifying whether a given program meets some specifications and failures to meet specification are generally communicated via counterexamples.

\section{Discussion and Future Directions}
The work discussed in this paper investigates the problem of generating explanations for unsolvability of a given planning problem. We also saw how the same methods apply when dealing with problems with plan constraints. In addition to extending these methods to more expressive domains, an interesting extension would be to try tackling cases where the current problem is solvable but all the solutions are too expensive. While this additional cost threshold could be seen as a constraint, the setting becomes a lot more interesting when the action costs are affected by the abstractions (c.f state dependent costs \cite{geisser2016abstractions}). With respect to contrastive explanations, this would correspond to cases where the alternative posed by the user is more expensive than the plan proposed by the robot.Finally, an obvious challenge to fully realize this method in practical scenarios is to develop methods to convert user questions to plan constraints. Methods like \cite{tenorth2010understanding} can be used to convert natural language statements to constraints like partial plans. Expert users can also directly write LTL and procedural programs as a way of interrogating the system.

\section*{Acknowledgments} This research is supported in part by the ONR grants N00014-16-1-2892, N00014-18-1-2442, N00014-18-1-2840, the AFOSR grant FA9550-18-1-0067, the NASA grant NNX17AD06G. and NSF grant 1844325.
\bibliographystyle{aaai}
\bibliography{Current_Draft}
\end{document}